\documentclass[final,5p,times,twocolumn]{elsarticle}
\usepackage{amssymb}
\usepackage{framed,multirow}
\usepackage{latexsym}
\usepackage{url}
\usepackage{xcolor}
\usepackage{hyperref}
\definecolor{newcolor}{rgb}{.8,.349,.1}
\usepackage[numbers]{natbib}
\usepackage{multirow}
\usepackage{algpseudocode}
\usepackage{amsmath}
\hypersetup{unicode=true}
\begin{document}


\begin{frontmatter}

\title{Supermarket Product Detection and Recognition: Utilizing Deep Learning with Rectified Imagery}


\author[1]{Mayank Sah\corref{cor1}}
\cortext[cor1]{Corresponding author: 
  email: mayank\_2221cs09@iitp.ac.in}
  
\author[1]{Jimson Mathew\corref{cor2}}
\cortext[cor2]{jimson@iitp.ac.in}

\address[1]{Department of Computer Science and Engineering, Indian Institute of Technology, Patna, India}


\begin{abstract}
   Product Identification has sprung up to become one of the most challenging problems in the automation of the retail industry. With the new industry 5.0 standards, automated inventory management, and catalog creation tasks are vitally important. Object identification models have emerged as a viable answer with their unprecedented identification and localization accuracy. However, the close-knit rack design of supermarkets generates the problem of angle variation in capturing images. The angle-variant densely packed images(a single image contains many objects) become overwhelming for these models alone. In this paper, we try to supplement object detection models with traditional Hough transform (HT) and homogeneous estimation concepts. We study the effect of rectified images using homography estimation and hough transform and their limitations on the problem of grocery identification. We make a case for creating a new dataset to test the effects of such rectification and produce analytical results on different scenarios of angle variation and object densities per image. Extensive experiments on different object detection models suggest that image rectification of angled images improves the detection accuracy of grocery products in images. The results also highlight the limitation of rectification on the angle of image capture and the object density of the image.
\end{abstract}


\end{frontmatter}

\section{INTRODUCTION}
\label{sec:introduction}
     \begin{figure*}[ht!]
        \centering
        \includegraphics[scale=.8]{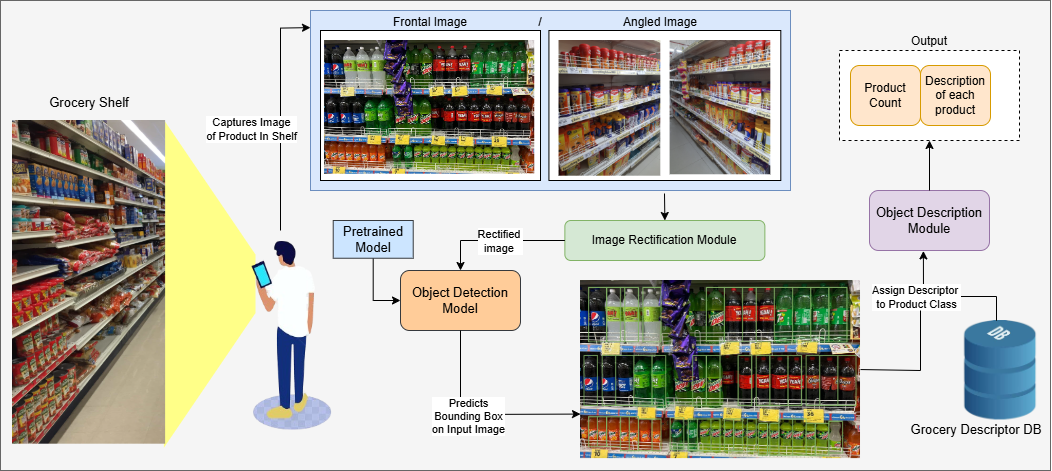}\\
        \caption{The illustration depicts the scheme of grocery identification. A user captures an image of the grocery shelf. The image is then passed through an object detection module that identifies the input image's products. The identified products are then passed through a description database that associates product descriptions with each product.}
        \label{Scheme}
    \end{figure*}

The transition toward Industry 5.0 has increased the need for automated inventory management, product identification, and catalog generation in retail environments. Although grocery-product identification is straightforward for humans, its automation remains challenging due to the large number of visually similar products, dense shelf arrangements, occlusion, and variations in imaging conditions. The scale of modern retail further emphasizes the need for reliable and scalable computer-vision-based solutions. According to Capital One Shopping Research \cite{retail_statistics}, global retail sales reached USD 29.3 trillion, highlighting the importance of efficient automation across both physical and online retail. Advances in computer vision have established object detection as an effective approach for automated product identification. Early feature based techniques \cite{HOG} were followed by deep learning-based detectors, including R-CNN \cite{RCNN}, SPPNet \cite{SPPNet}, YOLO \cite{YOLOv3,YOLOv5,YOLOv7,YOLOv8}, and RetinaNet \cite{RetinaNet}. Subsequent developments introduced anchor-free detectors, such as FCOS \cite{tian2020fcos}, and CenterNet \cite{duan2019centernet}, along with adaptive anchor-based approaches \cite{xie2023mutual,liu2024effective} and transformer-based architectures such as DETR \cite{DETR}. These methods have demonstrated strong performance across diverse applications, including crack detection \cite{crack} and food detection \cite{food}. However, their effectiveness in grocery-product identification can be limited by the characteristics of real-world supermarket scenes. Grocery shelves typically contain numerous closely packed and visually similar products, often captured from oblique viewpoints. Such viewpoints introduce perspective distortion, nonuniform object scales, and geometric deformation, making accurate localization and recognition particularly challenging in dense grocery scenes.

Our previous work, RectyNet \cite{rectynet}, demonstrated the potential of viewpoint correction for dense grocery-product detection using trainable rectification modules. However, incorporating additional trainable components increases model complexity. In this work, we instead investigate rectification as an independent preprocessing stage and evaluate its effect on single-stage and transformer-based object detectors. The captured image is first analyzed to estimate its dominant orientation, followed by homography-based perspective correction. The resulting rectified image is then provided to a pretrained object detector for product localization and recognition. To estimate the dominant orientation without introducing an additional learned module, we employ the Hough Transform (HT), a classical technique for detecting geometric structures in images. By mapping image-space lines into a parameter space, HT identifies the dominant shelf orientation from structural edges present in grocery-shelf images. Hassanein et al. \cite{HTTheory} provide a comprehensive review of Hough-based techniques and their applications in computer vision. The estimated orientation is then used to select an appropriate homography for correcting left- and right-oriented views. We initially explored this approach in our previous work \cite{rectification}; however, that study was limited in terms of experimental scope and analysis. In this work, we provide a more comprehensive evaluation of image rectification, investigating its effect across different viewpoints, object densities, and state-of-the-art object detection models.

Several datasets have been developed for grocery-product detection, including SKU110K \cite{SKU110K} and the Freiburg Groceries Dataset \cite{Freiburg}. However, these datasets do not explicitly focus on the effect of systematic camera viewpoint variation on dense product detection. Therefore, we use a subset of our previously curated Grocer-Help dataset \cite{gh}, containing frontal, left-oriented, and right-oriented grocery-shelf images, to systematically analyze the effects of viewpoint and perspective rectification.

The main contributions of this work are as follows:

\begin{enumerate}

\item A geometry-based rectification framework combining the Linear Hough Transform and homography estimation is developed to estimate dominant orientation and correct perspective distortion without adding trainable modules to the detector.

\item The effect of image rectification is systematically evaluated across single-stage and transformer-based object detectors and different camera orientations.

\item The influence of camera viewpoint and object density on rectification effectiveness is analyzed, including the limitations of rectification in dense grocery scenes.

\end{enumerate}

   \begin{figure*}[ht]
        \centering
        \includegraphics[scale=0.33]{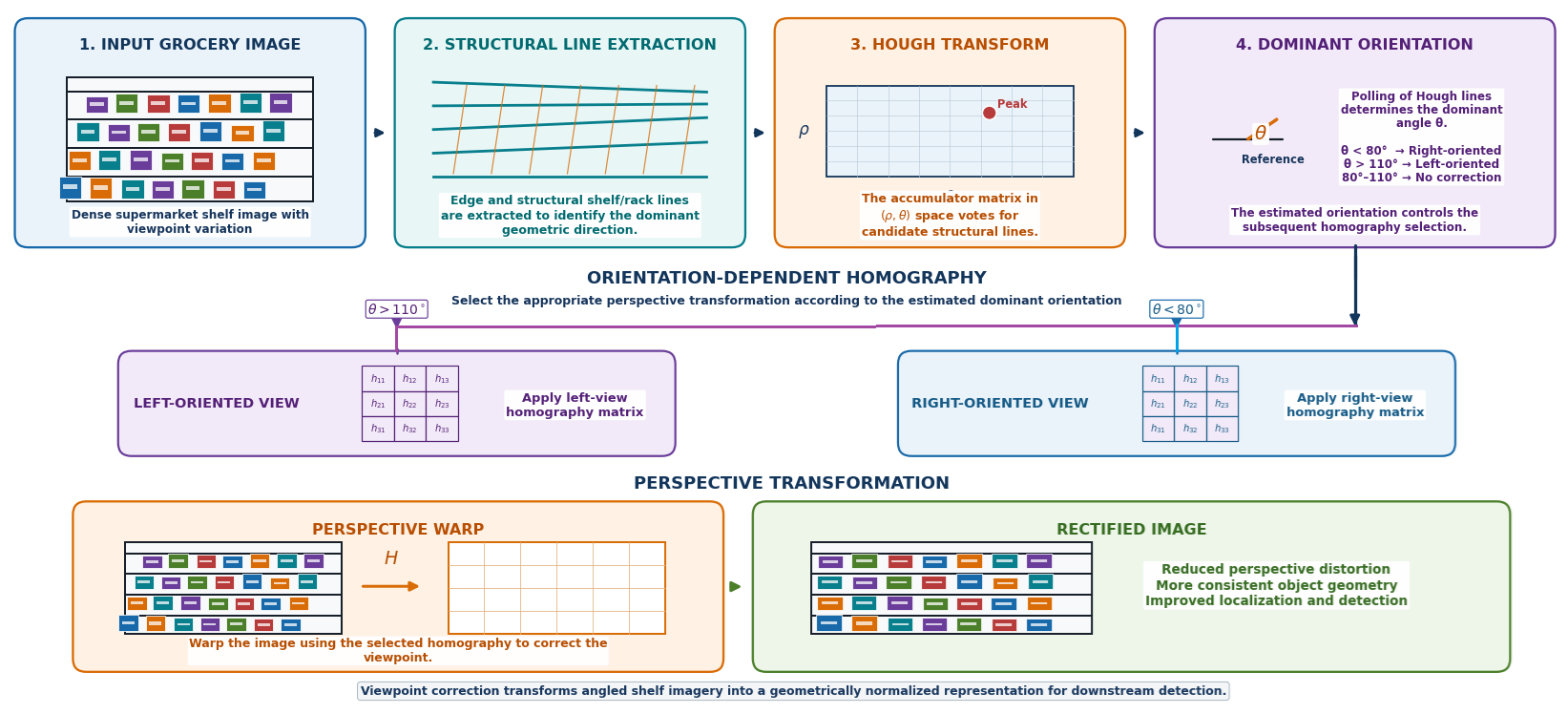}
        \caption{The figure shows the process of rectifying angled images. As input, two pairs of images (Right angled+front view, left angled+front view) are taken as input, and appropriate homography is estimated. Hough lines are generated on the test image, which is associated the image's alignment. Input image with appropriate homography generates a rectified image. }
        \label{methodology}
        \end{figure*}

\section{ METHODOLOGY}
\label{data}
    In this section, we provide a detailed description of the methodology followed. Fig. \ref{methodology} illustrates the process of image rectification in test images. The methodology deals with two main aspects, 
    \begin{enumerate}
        \item Angle identification using Hough transform.
        \item Homography estimation for Left and right aligned images.
    \end{enumerate}

    A test image is taken as input for the angle identification, and hough lines are detected in the test image.  The angle of this representative line decides the orientation of the image, whether the image is left-aligned or right-aligned. The image's alignment is decided by the camera's position that captures the image. For homography estimation, instances of known images are pushed in. Traditional feature-based methods are used in our work for ease of implementation. The homographies are estimated for two scenarios: left-aligned images named $H_L$ and right-aligned images named $H_R$. Finally, the test image is rectified based on the homography, and hough transform to provide a frontal view of an angled image. We have used a passive integration of Hough transform-assisted Homography estimation with the object detection models. The passive integration signifies that the homography estimation is done on two sets of representative images from the dataset before going for the object detection scheme. The two representative sets contain a pair of left-aligned images with its frontal image and a pair of right-aligned images with its frontal image. The homography matrices generated from the representative sets are maintained statically and remain the same for all the experiments. In the existing datasets and our dataset, the number of frontal view images heavily outnumbers the left-aligned and right-aligned images, thus providing skewed training for the object detection model. Thus, training the model to identify angled images correctly becomes more problematic. Using the image rectification module, we correct the angle variation in the angled images and warp the image into a frontal view. As a result, the rectified image is similar to the images used for training and thus improves the performance of the existing object detection algorithms. To summarize the methodology, we capture an image of the grocery shelf from our camera; the captured image is taken through an image rectification module, which provides a warped frontal image in case of an angled image or passes the image as it is when there is no significant angle deviation recorded. The warped images then travel through an object detection model and predict objects.
    
    For angle estimation in the test images, linear hough transform was used to generate hough lines using Python CV2 4.7.0 version with a canny edge detector and a constant threshold of 500. The angles of the generated hough lines depicted the alignment of the image. Homography was estimated by first resizing the input image to 640x640, followed by a SIFT feature generator to create key points in the aligned image and its frontal counterpart. The key points are then matched using a flann-based matcher algorithm to match the key points in the representative set. The matched key points are calculated using a RANSAC algorithm to calculate the fundamental matrix and inliers. This fundamental matrix is then used to rectify the images. The rectified image is then taken through an object detection model pipeline. Section 4 discusses the experimental results of the methodology in detail. We further divide this section into three subsections where the first subsection describes the dataset accumulation and annotation process. The second subsection describes the process of generating hough transforms from the dataset images for identifying image orientation, and the Third Subsection deals with applying appropriate rectification operations to generate the rectified image for testing.

\subsection{Dataset}

A subset of the Grocer-Help dataset is used to evaluate the proposed viewpoint estimation and perspective rectification methodology. The selected subset contains substantial variation in viewpoint, camera-to-rack distance, object density, and product appearance, making it suitable for evaluating rectification under realistic grocery-store conditions. The images include frontal, left-oriented, right-oriented, and extreme left/right-oriented views. Images were captured at varying distances from the product racks, resulting in both sparse and densely populated scenes, with dense images containing at least 35 objects. The captured racks contain up to five shelf rows, while extreme-view images typically cover two to four rows.

\begin{table}[t]
\centering
\caption{Comparison of representative grocery datasets with Grocer-Help.}
\label{tab:grocer_help_comparison}
\setlength{\tabcolsep}{2.5pt}
\renewcommand{\arraystretch}{1.15}
\begin{tabular}{lcccccc}
\hline
\textbf{Dataset} &
\textbf{DS} &
\textbf{MS} &
\textbf{GD} &
\textbf{BC} &
\textbf{FG} &
\textbf{RW} \\
\hline

Grozi-120
& $\times$ & $\times$ & $\times$ & \checkmark & \checkmark & \checkmark \\

Freiburg
& $\times$ & $\times$ & $\times$ & \checkmark & \checkmark & \checkmark \\

RPC
& \checkmark & $\times$ & $\times$ & \checkmark & \checkmark & \checkmark \\

SKU110K
& \checkmark & \checkmark & \checkmark & $\times$ & $\times$ & \checkmark \\

\textbf{Grocer-Help}
& \textbf{\checkmark}
& \textbf{\checkmark}
& \textbf{\checkmark}
& \textbf{\checkmark}
& \textbf{\checkmark}
& \textbf{\checkmark} \\

\hline
\end{tabular}

\vspace{2pt}
\begin{flushleft}
\scriptsize
DS: Dense Scenes; MS: Multi-store Collection; 
GD: Geographic Diversity; BC: Brand-based Classes; 
FG: Fine-grained Products; RW: Real-world Retail Setting.
\end{flushleft}
\end{table}

\begin{figure*}[ht]
        \centering
        \includegraphics[scale=0.47]{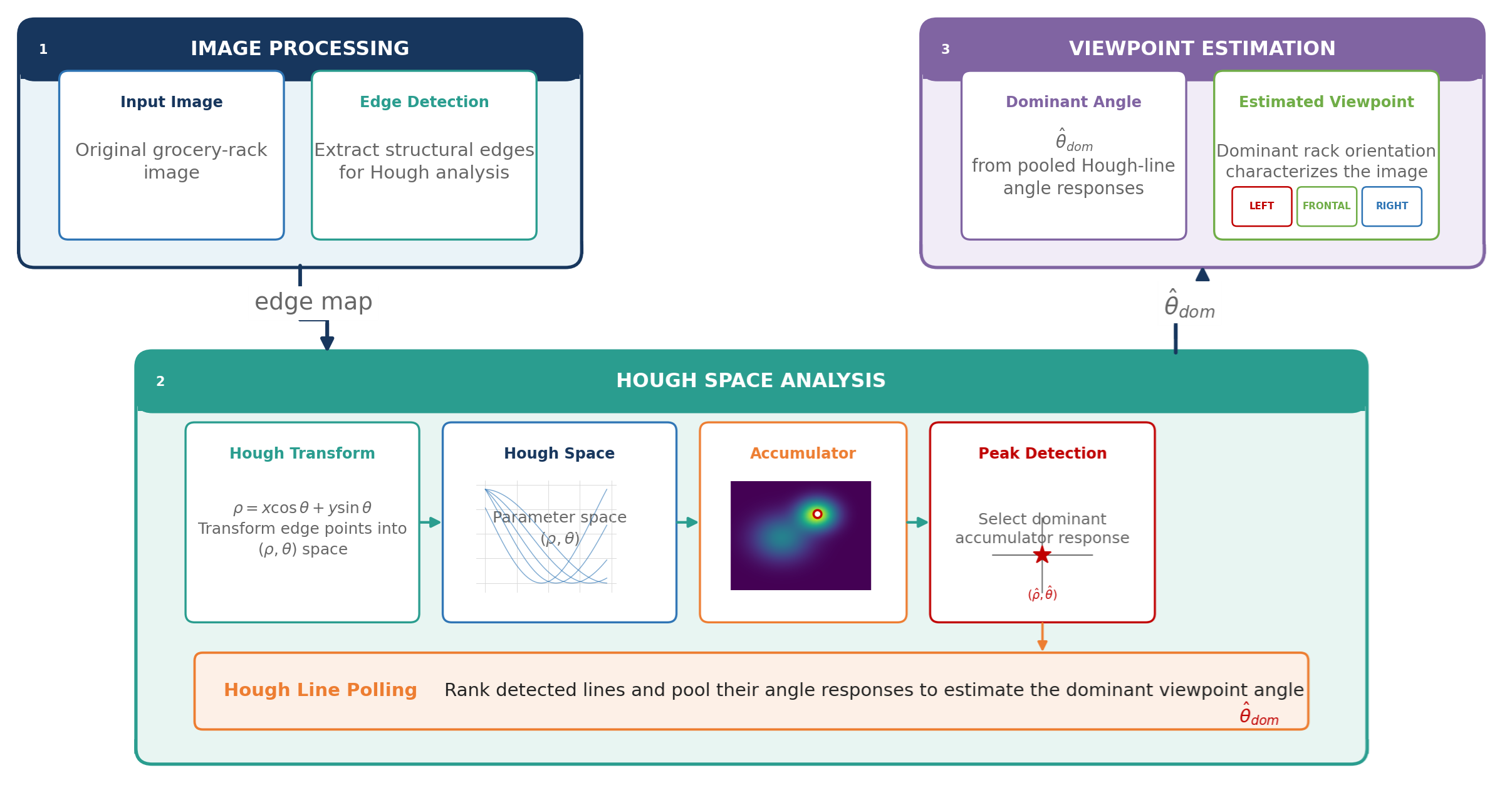}
        \caption{The image illustrates the procedure for generating Hough Lines and Image Instances of Hough Transform on left-aligned and right-aligned images}
        \label{hough transform}
        \end{figure*}
        
The subset contains 1946 images, with 1310 images used for training and 636 for testing. Training-time augmentation using contrast, saturation, blur, hue, and flipping increases the training set to 7784 images, of which 5240 images constitute the augmented training set. The evaluation set comprises 2544 images, including 944 frontal, 800 right-oriented, and 800 left-oriented images. The frontal images include both close- and long-range views, providing variation in object scale and density. A total of 176 product classes are considered.

To ensure implementation consistency, all images are resized to $640\times640\times3$. No additional preprocessing is applied. The images are manually annotated with bounding boxes, and class assignments follow the product grouping of the Grocer-Help dataset, where visually similar products may share a class while visually distinct products from the same brand are retained as separate classes.

\subsection{Using Hough Transform to identify image orientation}
\label{data1} 
  
    The Hough Transform (HT) generally transforms an image from image space to parameter space. This conversion to parameter space leads to the problem of curve detection in the image and the peak detection problem in parameter space. In general terms, if we consider a line in image space, then this line converts to a point in the parameter space. Figure \ref{hough transform} showcases the procedure for calculating angles for different angle-aligned images. Our implementation uses a Probabilistic Hough Transform instead of a Simple Hough Transform. The input image is first transformed into a greyscale image, and an accumulation matrix is calculated from the greyscale image.  The voting mechanism finds the points where a high number of lines intersect. The accumulation matrix then uses a voting mechanism to highlight the key points in the Euclidean space and transform them into a parameter space. The hough lines are then generated from the parameter space. The parameter space also provides the $\theta$ and $\rho$ value for the test image, providing the test image's polarity.
    
    In our method, we use the Hough transform in conjunction with the edge detection module, namely the Canny Edge Detection module, and thus, deal with straight-line parameterization.

    The equation of a line is written as:
    \begin{equation}
        y=mx+c
    \end{equation}

    Considering this eqn. (1) in the image space, we define the image in terms of \textit{(x,y)} whereas in transformed space, we define the image in terms of \textit{(m,c)}. Thus transforming the image parameters. A further parameterization of the equation is done as m and c can be infinite; thus, to curtail this down, we get a parameterized equation as:
    \begin{equation}
        y Cos(\theta) = x Sin(\theta) + \rho
    \end{equation}
    where, $0 \leq \theta \leq \pi$.

    Based on this concept of HT, we input the grocery images and generate the hough lines in the input image, as shown in the figure.\ref{hough transform}.
 
    The lines generated as the HT output decide the image's orientation.
    
    \begin{algorithmic}
        \If{$\theta < 85$} 
        \State $Image$ $is$  $"LEFT"$ $Oriented$
        \EndIf
     \end{algorithmic}
    \begin{algorithmic}
        \If{$\theta > 95$} 
        \State $Image$ $is$ $"RIGHT"$ $Oriented$
        \EndIf
    \end{algorithmic}

\begin{figure*}[ht]
        \centering
        \includegraphics[scale=0.48]{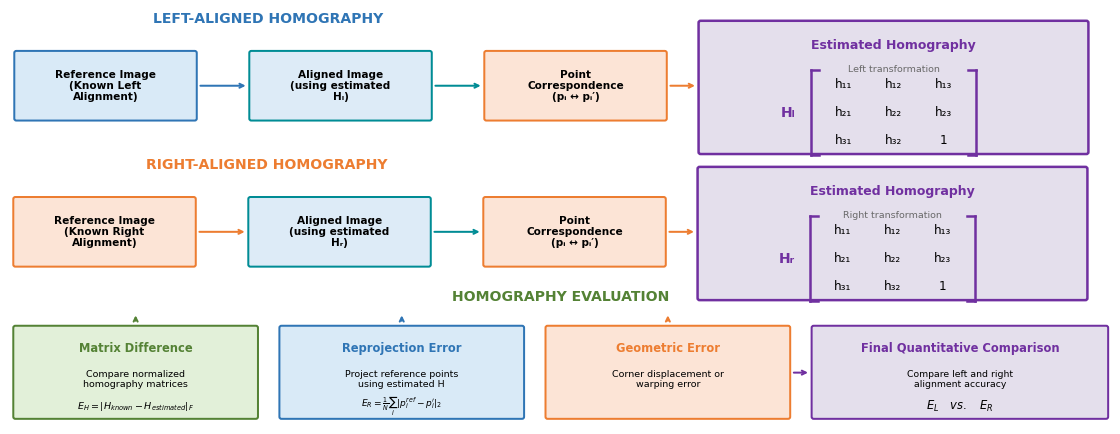}
        \caption{The figure illustrates the effect of Image rectification on Left and Right Aligned Images using respective aligned Homography}
        \label{Homography}
        \end{figure*}

As the deviation in line angle is very low on the shelf images, thus the range of  $\theta$ is kept very low. Fig. \ref{hough transform} showcases the low deviation of angles on the right-aligned and left-aligned images' output images. The hyperparameters (Accumulator threshold parameter) in the Hough line generator are maintained such that the shelf display angles take priority over the edges detectable from the product boundaries, as the product boundaries may cause erroneous image angle detection.

   \subsection{Homography Estimation and Image Rectification}

In our dataset, we consciously chose to take images in right-aligned and left-aligned scenes. These images capture the same data with different viewpoints and thus contain a different feature set. As we have also emphasized variation in camera angles, there is variation among the images of the same orientation. This variation causes multiple homography matrices for different angles. We only use a single-layer homography here, so we take a representative left and right homography matrix at 40 to 60 degrees in each alignment. We derive two homography matrices for the two viewpoints to analyze the best possible object detection scenario.
\begin{enumerate} 

\item The first homography matrix takes the input images that are right aligned and creates a homography matrix according to its perspective straight view. The right-aligned homography matrix is denoted as $H_R$. 

\item The second homography matrix takes the input images that are left aligned and creates the homography matrix with its perspective straight view. The left-aligned homography matrix is denoted as $H_L$
\end{enumerate}

The rectification of the input test image is done by a vector dot product of the input image with the appropriate homography matrix. The output is a perspective projection of an angled image in frontal view.
  
Each viewpoint in the image space can be seen as a matrix. Suppose we take our aligned image as a matrix A. We are required to transform this matrix A to another matrix B, which is the frontal view of the image. Thus, this operation is possible using the following equation \cite{HomographyReview}. If the image is left aligned:

\begin{equation}
    B = A.H_L 
\end{equation}

If the image is right aligned:
\begin{equation}
    B = A.H_R
\end{equation}
where, H is the homography matrix \cite{HomographyReview} such that,

\begin{equation}
    H = \begin{vmatrix}
h_{11} & h_{12} & h_{13}\\
h_{21} & h_{22} & h_{23}\\
h_{31} & h_{32} & h_{33}
\end{vmatrix}
\end{equation}
where,
\begin{equation}
    h_{11}^2 + h_{12}^2 + h_{13}^2 + h_{21}^2 + h_{22}^2 + h_{23}^2 + h_{31}^2 + h_{32}^2 + h_{33}^2 = 1
\end{equation}

where, $[h_{11}, h_{12}, h_{21}, h_{22}]$ represent the affine transformation, $[h_{13}, h_{23}]$ represents the translation transformation and $[h_{31}, h_{32}]$ represents the perspective transformation values. $h_{33}$ is normalized to 1 so that the homography matrix has a degree of freedom to be 8. The homography estimation module determines the $H_L$ and $H_R$ values. 'B' is the input image, and 'A' is the rectified image, transforming the angled image into a rectified frontal image. The values of $[h_{11}, h_{12}, h_{21}, h_{22}, h_{13}, h_{23},h_{31}, h_{32}]$ are calculated by matching the key-points of the angled image and its equivalent frontal image by a flann based matcher, and then running RANSAC algorithm on the identified key-points. 

A wise procedure was maintained to calculate the homography matrices:
\begin{enumerate}
    \item \textbf{Input:} Aligned image and Normal Image of some similarity
    \item Generate Keypoint Features and Descriptors for both the input images
    \item Match Descriptor of the aligned image with normal image and generate point mappings
    \item Choose Point maps and compute the homography using the RANSAC algorithm
    \item \textbf{Output:} Homography matrix H
\end{enumerate}

\setlength{\arrayrulewidth}{.5mm}
    \setlength{\tabcolsep}{5pt}
    \renewcommand{\arraystretch}{1.1}
    {

        \begin{table*}[htb!]
        \centering
        \footnotesize
        \label{table:1}

        \begin{tabular} {p{.15\linewidth} p{.07\linewidth} p{.07\linewidth} p{.08\linewidth} p{0.08\linewidth} p{.08\linewidth} p{.08\linewidth} p{.08\linewidth} p{.08\linewidth}} \hline
             \textbf{\centering{Dataset}}   & \textbf{\centering{Images}}   &    \textbf{Classes} & \textbf{YOLOv3} & \textbf{YOLOv5m} & \textbf{YOLOv7} & \textbf{YOLOv8m} & \textbf{YOLOv9} & \textbf{RetinaNet}\\ 
        \hline
            GroZi-120\cite{Grozi120} & 11870 & 120 & 24.1 & 23.9& 25.61 & 25.2 &26.3 & 24.1 \\
            Webmarket\cite{Webmarket} & 3153 & 200 & 29.4 &29.2 & 31.2 & 31.8 & 32.1  & 31.3 \\
            GroZi-3.2K\cite{Grozi3.2} & 8350 & 80 & 27.1 & 26.1 & 27.5 & 26.7 & 29.1 & 26.6 \\
            FreiBurg\cite{Freiburg} & 4947 & 25 & 88.1 & 73 & 79.35 & 92.4 & 93.5 & 78.85\\
            SKU110K\cite{SKU110K} & 11762 & 1 & 91.7 & 89 & 88.49 & 90.9 & 92.2 & 90.0\\    
             \hline
                 
        \end{tabular}
        \caption{mAP of SOTA models on existing Grocery Dataset}
        \label{tab3}
       
        \end{table*}
 Fig. \ref{Homography} showcases the difference in the product visibility in an angled image and its rectified image. The image rectification module increases the visibility of grocery products. However, image rectification may lead to some loss of information from the non-rectified images. The steeper the viewing angle, the more the loss of information in rectification loss. Even though image rectification leads to information loss, it propels the product's identification accuracy on a shelf. The rectification results also highlight a constraint that the images should not be very densely packed, as the information loss in such images outweighs the increase in identification accuracy.

\section{EXPERIMENTAL SCENARIO AND ANALYSIS}
\label{R}
 
    This section explains the experimental setup referred to for training and testing images. We also provide a qualitative analysis of training and testing our dataset of state-of-the-art methods, namely YOLOv3\cite{YOLOv3}, YOLOv5\cite{YOLOv5}, YOLOv7\cite{YOLOv7} and YOLOv8\cite{YOLOv8} and RetinaNet\cite{RetinaNet}. Table \ref{results} highlights the results achieved on our dataset on state-of-the-art models.  

   \begin{figure}[ht!]
        \centering
        \includegraphics[scale=.85]{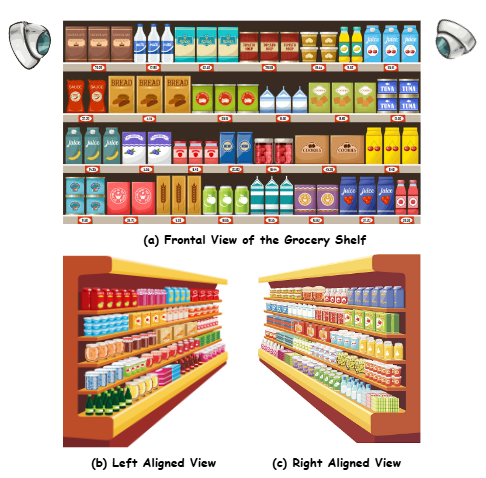}\\
        \caption{The Figure highlights the camera angle's effect on the image's view. Image (a) showcases a frontal view of a grocery rack, (b) Depicts the view of the rack when the image is taken from a left-aligned camera, and (c) Depicts the view of the rack when the image is taken from a right aligned camera.}
        \label{scenario}
    \end{figure}

\subsection{Experimental scenario}

    We captured images from a single grocery store to test the effects of image rectification on object identification. The selection of grocery stores was made such that there was variation in product brands in every rack to accommodate more diversity in image features. The camera angles were also maintained to capture a more holistic view of the racks. Fig. \ref{scenario} showcases the camera angle's effect on the image's view.

    A steep camera angle would lead to an image requiring more rectification. In our works, we have considered only a single-layer rectification; thus, the steep view of images leads to higher information loss. The angled images are used for testing, whereas the training images are a mix of frontal and angled images. The experiments were carried out on a GeForce GTX1080Ti GPU with 100 epochs. The training set was maintained at 4800 images with over 60,000 objects. For uniformity of experimentation throughout models, hyper-parameters like Input image size, learning rate, weight decay, and momentum were maintained and kept at 640, 0.01, 0.0005, and 0.937, respectively. The objects per image varied from 2 to 55. For testing, 200 right-aligned images, 200 left-aligned images, and 136 front-view images were maintained, and the camera angles for angled images varied from 15 to 80 degrees at each viewing angle. We have only used one-stage object detection models for experimentation as they are much faster than traditional two-stage models. We have also used a transformer model as the parameters, and floating point operations almost matched the one-stage detector methods. 

\begin{figure*}[ht!]
        \centering
        \includegraphics[scale=.44]{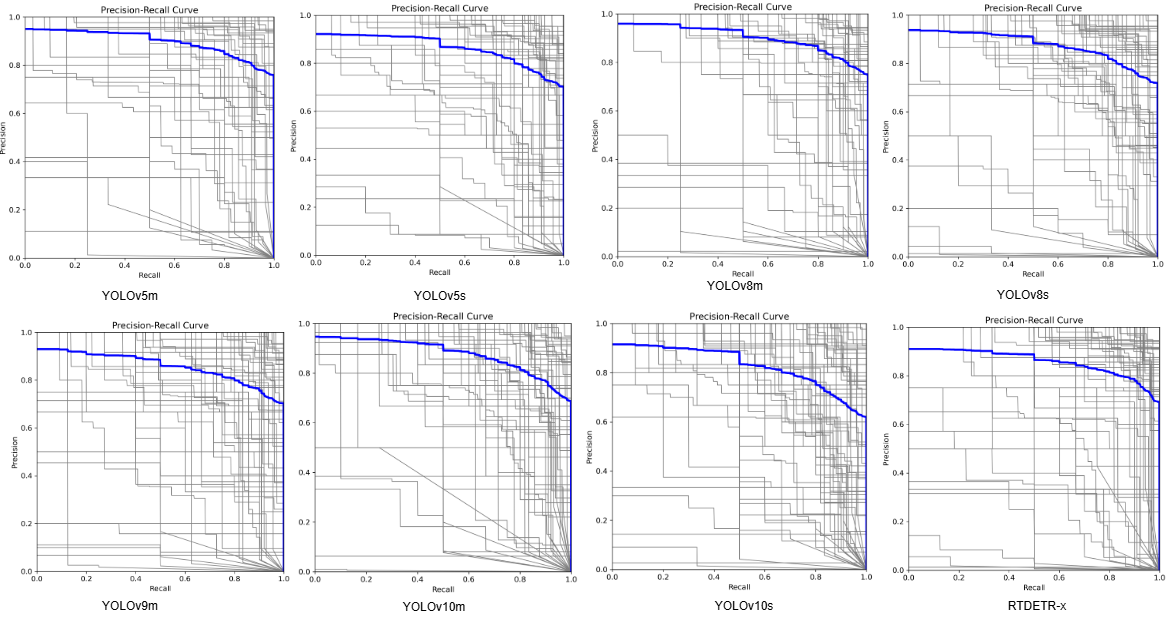}\\
        \caption{The Figure highlights the Precision-Recall curve of best-performing algorithms on our dataset}
        \label{PR}
    \end{figure*}

\setlength{\arrayrulewidth}{.45mm}
    \setlength{\tabcolsep}{2 pt}
    \renewcommand{\arraystretch}{1}
    {
 \begin{table*}[htpb!]
    \footnotesize
    \centering
    \setlength{\tabcolsep}{6pt} 
    \renewcommand{\arraystretch}{1.1}
    \begin{tabular}{p{.1\linewidth} p{.07\linewidth} p{.09\linewidth} p{.06\linewidth} p{0.06\linewidth} p{.06\linewidth} p{.06\linewidth} p{.06\linewidth} p{.06\linewidth} p{.06\linewidth} p{.06\linewidth}}
        \hline
        \multirow{3}{*}{ \textbf{Model}} & \multirow{3}{*}{ \textbf{FLOPS}}    & \multirow{3}{*}{ \textbf{Parameters}} & \multicolumn{6}{c}{\textbf{Non Rectified Images}} & \multicolumn{2}{c}{\textbf{Rectified Images}}  \\ 
        
        & & & \multicolumn{2}{c}{\textbf{Frontal View}} & \multicolumn{2}{c}{\textbf{Right Angled}} & \multicolumn{2}{c}{\textbf{Left Angled}} & \multirow{2}{*}{ \textbf{mAP 0.5}} & \multirow{2}{*}{ \textbf{mAP 0.9}}\\ 

        & & & \textbf{mAP 0.5} & \textbf{mAP 0.9} & \textbf{mAP 0.5} & \textbf{mAP 0.9} & \textbf{mAP 0.5} & \textbf{mAP 0.9} &  & \\
        
    \hline

        \textit{YOLO v3}   & 283.7 G & 103.83 M & 87.6 & 73.8 &  52.23 & 38.67 & 51.1 & 39.7 & 54.79 & 40.12\\
        
        \textit{YOLO v5n}   & 8.8 G & 2.86 M & 60.1 & 46.2 &  49.12 & 40.1 & 49.4 & 41.38 & 56.2 & 45.1\\

        \textit{YOLO v5s}   & 24.4 G & 9.19 M & 83.9 & 65.5 &  52.3 & 41.2 & 53.1 & 41.7 & 57.2 & 45.2\\

        \textit{YOLO v5m}   & 64.9 G & 25.17 M & 88.3 & 71.4 &  54.1 & 41.6 & 55.1 & 45.2 & 58.6 & 43.9\\

        \textit{YOLO v7}   & 108.1 G & 38.14 M & 74.34 & 62.7 & 52.3 & 41.4 & 50.5 & 38.7 & 56.2 & 47.3\\

        \textit{YOLO v7-tiny}   & 14.7 G & 6.48 M & 62.1 & 50.6 &  40.2 & 28.1 & 41.2 & 29.1 & 45.2 & 36.0\\
        
        \textit{YOLO v8n}   & 7.4 G & 2.77 M & 42.2 & 33.9 &  34.3 & 28.1 & 34.4 & 28.1 & 36.1 & 28.8\\

        \textit{YOLO v8s}   & 29.0 G & 11.2 M & 82.0 & 65.30 &  53.1 & 42.12 & 53.9 & 43.5 & 56.4 & 47.8\\

        \textit{YOLO v8m}   & 79.6 G & 25.96 M & 85.77 & 70.24 &  55.1 & 42.45 & 56.8 & 45.1 & 59.5 & 49.1\\
        
        \textit{YOLO v9t}   & \textbf{7.1 G} & 1.8 M & 37.8 &  30.5 &  29.7 & 22.1  & 30.1 & 22.4 & 33.2 & 25.1\\

        \textit{YOLO v9s}   & 23.1 G & 6.3 M & 64.9 &  54.7 &  52.8 & 44.8  & 52.1 & 46.2 & 57.3 & 48.6\\

        \textit{YOLO v9m}   & 61.8 G & 16.85 M & 83.4 &  68.7 &  \textbf{57.8} & 45.6  & \textbf{58.1} & 46.2 & 60.1 & 49.7\\
        
        \textit{YOLO v10n}   & 9.3 G & 2.88 M & 51.2 &  41.4 &  40.0 & 37.8  & 40.2 & 37.9 & 44.5 & 39.2\\

        \textit{YOLO v10s}   & 25.5 G & 8.2 M & 80.3 &  65.7 &  54.2 & 46.6  & 55.0 & 47.1 & 60.1 & 50.0\\
        
        \textit{YOLO v10m}   & \textbf{65.1 G} & 16.69 M & \textbf{86.7} &  71.4 &  57.1 & 44.9  & 57.8 & 46.1 & \textbf{60.8} & 50.7\\
        
        \textit{RTDETR-l}   & 108.8 G & 33.17 M & 50.4 &  39.8 &  40.2 & 30.3  & 41.0 & 30.9 & 42.5 & 31.1\\
        \textit{RTDETR-x}   & 233.1 G & 67.66 M & 79.2  & 65.1  & 52.5  & 43.1 & 53.2 & 43.3 & 57.2 & 47.1\\
        
        \textit{Retina-Net}   & 60 G &  34 M & 76.9 & - & 56.3  & - & 56.9 & - & 59.8 & -\\

        \hline
    \end{tabular}
    \caption{ The table summarises the accuracy of different Object detection models on our dataset with a frontal view, angled view, and rectified images with mAP 0.5 and mAP 0.9 as the evaluation metric (Here G stands for Giga and M stands for Million)}
    \label{results}
    \end{table*}
    }     
    \subsection{Experimental Analysis}

     The existing datasets mostly contain images in frontal view; thus, the rectification module cannot be tested on these existing datasets. The results in table \ref{tab3} showcase the accuracy of the models we use in our analysis of the existing datasets. The same models and their variants are applied to our dataset with the same hyperparameters to analyze the impact of image rectification on product identification.

\setlength{\arrayrulewidth}{.5mm}
    \setlength{\tabcolsep}{5pt}
    \renewcommand{\arraystretch}{1.1}
    {
\begin{table}[ht]
    \footnotesize
    \centering
    \begin{tabular}{ccccc }
        \hline
        \multirow{2}{*}{\textbf{Angle}} & \multicolumn{2}{c}{\textbf{O/I} $\leq$ \textbf{10}} & \multicolumn{2}{c}{\textbf{O/I} $\geq$ \textbf{11}} \\
        & \textbf{mAP 0.5} & \textbf{mAP 0.5 (R.I)} & \textbf{mAP 0.5} & \textbf{mAP 0.5 (R.I)}\\
       
    \hline

         10 - 30 & 41.2 & 45.5 & 14.3 & 13.9 \\
        30 - 60 & 74.8 & 80.2 & 34.2 & 36.1 \\
         60 - 80 & 83.5 & 84.0 & 56.2 & 57.1  \\
         110 - 130 & 83.1 & 83.2 & 57.1 & 56.3 \\
        130 - 150 & 71.6 & 75.7 & 36.1 & 36.8\\
        150 - 170 & 40.08 & 42.2 & 11.01 & 11.00\\
        \hline
    \end{tabular}
    \caption{The table highlights the effect of angle variation in mAP 0.5 on densely packed and sparsely packed images. Here O/I is Objects per Image and R.I is Rectified Image }
    \label{tabscenario1}
    \end{table}
} 
  The experiments carried out with our dataset on state-of-the-art single-stage detector models are tabulated in Table \ref{results}. The analysis is done in four vertical segments. The first segment elucidates the results on only frontal view images with no angle deviation. This situation showcases the best results over all the models used. The second vertical segment contains all images that are right aligned. The right-aligned images also vary in the image capture angle, and the results show that the identification accuracy drops significantly in such a scenario. The third vertical segment is similar, but the images are all left-aligned with angle variation. This variation also showcases a dip in detection accuracy but is identical to that of right-angled images. All the cases discussed are non-rectified, i.e., our image rectification module has not yet worked in any of the scenarios. Our goal is to convert the angled images into frontal view, so we only take the angled images for rectification. The last vertical segment showcases the results after rectifying the angled images.  

  \begin{figure*}[ht!]
        \centering
        \includegraphics[scale=.52]{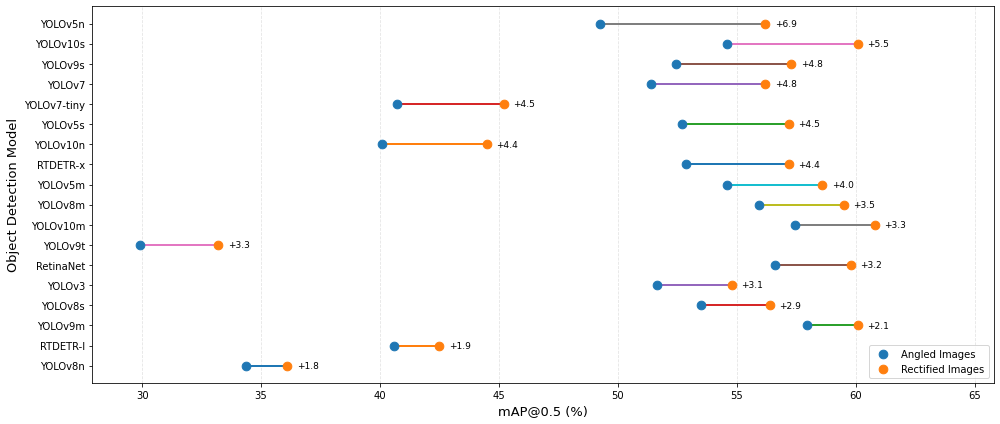}\\
        \caption{The Figure highlights the gain perceived in all the existing algorithms when run with rectified images in comparison to the angled images on Grocer-Help dataset}
        \label{comp}
    \end{figure*}
    
  Table \ref{results} illustrates that all the models perform the same despite the deviation in accuracy. All models work best with non-rectified frontal images and then showcase a dip when dealing with angled photos. The performance improves for all the models as we rectify the angled pictures. The best mAP 0.5 (mean average precision) in the case of frontal view image showcases the accuracy of identifying the correct grocery product is a decent 86.7, but if we take any pictures with specific alignment due to camera positioning, the MAP 0.5 drops from 86.7 to the range of 56 to 58, which is a significant drop in the accuracy of correct identification. The accuracy in rectified images is somewhat higher than the angled images at 60.8. The numbers are high but do not match the frontal view scenario because some loss in object boundaries is also encountered due to warping. Figure \ref{comp} highlights the gain every algorithm showcases when the angle information is truncated from the image.

The results published in table \ref{tabscenario1} showcase the effect of change in camera positioning on MAP when the O/I (Objects per Image) is maintained less than or equal to 10. In such a scenario, the best results are achieved when the camera position is maintained at an angle of 60 to 130 degrees. However, it is also worth noting that the best possible effect of image rectification is perceived when the camera angles are maintained at an angle of 30 to 60 degrees from either side. The same effect is observed when the O/I is maintained at greater than 10. It is noteworthy that the MAP drops considerably when the O/I increases. So, a densely packed image with a very steep inclination of 10 to 30 degrees shows inferior accuracy in grocery detection. The results of table \ref{tabscenario1} showcase that the best effect of rectification is achieved when the camera angle is maintained at 30 to 60 degrees. Thus, to study the impact of O/I specifically, we test 60 images when the camera angle is left aligned at 30 to 60 degrees and 58 images when the camera angle is right aligned at 30 to 60 degrees.
  \setlength{\arrayrulewidth}{.5mm}
    \setlength{\tabcolsep}{5pt}
    \renewcommand{\arraystretch}{1.1}
    {
\begin{table}[ht]
    \footnotesize
    \centering
    \begin{tabular}{ccccc}
        \hline
        \multirow{2}{*}{\textbf{O/I}} & \multicolumn{2}{c}{\textbf{30} $\leq$  \textbf{right\_alignment} $\leq$ \textbf{60}} & \multicolumn{2}{c}{\textbf{30} $\leq$  \textbf{left\_alignment} $\leq$ \textbf{60}} \\
       & \textbf{mAP 0.5} & \textbf{mAP 0.5 (R.I)} & \textbf{mAP 0.5} & \textbf{mAP 0.5 (R.I)}\\
       
    \hline

         less than 5 & 89.8 &  89.9 & 88.4 & 89.0\\ 
        5 - 10 &  85.3 &  86.4 & 84.1 & 84.9 \\
        10 - 20 & 79.2 & 83.1 & 78.8 & 83.1 \\
         20 - 40 &  49.1 & 50.0  & 48.2 & 50.0\\
        more than 40 & 18.2 & 15.1 & 19.1 & 18.2\\
       
         \hline
    \end{tabular}
    \caption{The table highlights the effect of Objects per Image (O/I) with the viewing angle restricted to 30$\leq  \theta\leq$ 60. $R.I stand for Rectified Image$}
    \label{tabscenario3}
    \end{table}
}

The impact of O/I is clearly understood from the results achieved in table \ref{tabscenario3}, which signifies that it is tough to identify angled images in a densely packed scenario. The O/I, when less than 5, showcases the highest identification accuracy; however, the effect of rectification in such a scenario is minimal as the objects are visible in the original image. Thus, the O/I range of 10-20 shows a considerable change in improving identification accuracy due to rectification. Therefore, image rectification works best when the O/I is kept at 10-20 and $\theta$ is maintained such that $30 \leq \theta \leq 60$ on either side.

\section{LIMITATIONS AND FUTURE WORK}
    
   Although image rectification improves detection performance for angled images, its effectiveness is subject to several limitations:

\begin{enumerate}

\item \textbf{Loss of image content in dense scenes:} For densely packed images captured at steep angles, rectification may remove objects near the image boundaries or introduce distorted and noisy regions. In extreme cases, the rectified image may contain only approximately 20\% of the original image content. This occurs primarily because a fixed homography produces a large geometric transformation for steep viewpoints. A gradual homography transformation could reduce this loss of information.

\item \textbf{Dependence on visible rack boundaries:} The Hough Transform performs reliably when shelf or rack boundaries are clearly visible. When these structural boundaries are absent, the detected lines may instead correspond to product edges, resulting in an inaccurate orientation estimate. This limitation is particularly evident in closely captured images or racks with minimal visible structural boundaries. Incorporating additional structural cues, such as vacant shelf regions, could improve orientation estimation when rack boundaries are unavailable.

\item \textbf{Limited correction for extreme viewpoints:} At very steep viewing angles, a single global homography may not adequately transform the image into a frontal view, resulting in partial or distorted rectification. The current approach uses fixed homography matrices for left- and right-oriented images irrespective of the magnitude of the viewpoint angle. A viewpoint-dependent or gradually varying homography could provide more effective correction across a wider range of angles.

\item \textbf{Single-rack constraint:} The current rectification framework assumes that the input image contains a single dominant rack orientation. Images containing multiple racks with different orientations may lead to ambiguous line detection and incorrect orientation estimation.

\end{enumerate}

These limitations indicate several directions for future research. A deep Hough Transform integrated into the detection pipeline could enable joint estimation of image orientation and object detection while reducing dependence on manually visible rack boundaries. For steep viewpoints, progressive or multi-stage homography correction could minimize information loss and distortion. In addition, lightweight and quantized detection models can be investigated to develop computationally efficient rectification-based detection systems suitable for practical retail environments.

\section{CONCLUSION}

    In this paper, we propose a dataset that contains standard authentic life images of grocery stores and their available products. We also specified some pictures with an angled or skewed view of the grocery products on shelves. We then introduced the idea of using Hough Transform to generate Hough lines in the angled images, providing us with the skewness of the objects in the image. The hough lines generated identify all test images as left or right-aligned. The aligned images are then rectified using the respective view homography matrix. As far as we know, image rectification using Hough transform-assisted Homography is the novelty we have proposed in our work, and no work has been carried out in this domain. The experiments carried out in this paper help establish the positive role image rectification plays in camera variation in grocery identification.  

     The results generated on single-stage object detection algorithms showcase that image rectification improves product identification if the images contain 5 to 10 objects. With the photos getting dense, the accuracy of prediction drops. In the future, we plan to solve the problem of applying hough transform in densely packed images to make this work more diverse, mimicking real-time conditions.

    
    Moreover, this paper deals with the problem of grocery identification in a real-life scenario where images may be from far away, from near, or from a random angle. Also, it may be a slight blur due to motion. This is quite in contrast to the existing datasets considered for experimental setups. We aim to extend the dataset to encapsulate more images per class and more classes available in the Indian market. In our future endeavour, we plan to implement object-wise regional attention to identify the products on the run. We plan to explore the domain of catalogue creation directly from images, which automates the grocery cataloguing process.

 \bibliographystyle{ieeetr}


 \bibliography{reference_new}

\end{document}